\pdfoutput=1

\documentclass[11pt]{article}

\usepackage[final]{acl}

\usepackage{times}
\usepackage{latexsym}

\usepackage[T1]{fontenc}

\usepackage[utf8]{inputenc}

\usepackage{microtype}

\usepackage{inconsolata}

\usepackage{graphicx}

\usepackage{amsmath,amsfonts,bm}

\def\eqref#1{equation~\ref{#1}}

\def\1{\bm{1}}

\def\vd{{\bm{d}}}

\def\vq{{\bm{q}}}

\def\vs{{\bm{s}}}

\DeclareMathAlphabet{\mathsfit}{\encodingdefault}{\sfdefault}{m}{sl}
\SetMathAlphabet{\mathsfit}{bold}{\encodingdefault}{\sfdefault}{bx}{n}

\usepackage{algorithm}
\usepackage{algorithmic}
\usepackage{setspace}
\usepackage{comment}
\usepackage{booktabs}
\usepackage{hyperref}
\usepackage{icomma}
\usepackage{url}

\usepackage[nameinlink]{cleveref}

\usepackage{longtable}
\usepackage{xcolor}
\usepackage{colortbl}
\usepackage{multicol}
\usepackage{multirow}
\usepackage{makecell}
\usepackage{amsmath, amssymb}
\usepackage{mathtools}
\usepackage{enumitem}
\usepackage{subcaption}
\usepackage{float}
\usepackage{graphicx}
\usepackage{caption} 
\usepackage{upgreek}
\usepackage{seqsplit}
\usepackage{color,soul}
\usepackage{arydshln}
\usepackage{xspace}
\usepackage{adjustbox}
\usepackage{enumitem}
\usepackage{booktabs} 
\usepackage[most]{tcolorbox}

\definecolor{Red}{rgb}{0.6,0,0}
\definecolor{Blue}{rgb}{0,0,0.8}
\definecolor{Green}{rgb}{0,0.4,0.7}
\definecolor{mountainmeadow}{rgb}{0.19, 0.73, 0.56}
\definecolor{crimson}{rgb}{0.86, 0.08, 0.24}
\definecolor{darkblue}{rgb}{0.0, 0.0, 0.55}

\definecolor{gg}{HTML}{E0FEFE}
\definecolor{gray}{RGB}{236, 236, 236}

\hypersetup{
    colorlinks = true,
    citecolor = mountainmeadow,
    linkcolor = crimson,
    linktoc = all,
    urlcolor = darkblue,
}

\newcommand{\highlight}[1]{{\color{crimson}{#1}}}

\newcommand{\ours}{IDentIfy\xspace}

\title{Unified Multimodal Interleaved Document Representation for Retrieval}

\author{
    Jaewoo Lee$^{1*}$ \; 
    Joonho Ko$^{2*}$ \;
    Jinheon Baek$^{2*}$ \;
    Soyeong Jeong$^{2}$ \;
    Sung Ju Hwang$^{2,3}$ \\
    University of North Carolina Chapel Hill$^{1}$ \; KAIST$^{2}$ \; DeepAuto$^{3}$ \\
    \texttt{jwoolee@cs.unc.edu, \{joonho.ko, jinheon.baek, sungju.hwang\}@kaist.ac.kr}
}

\begin{document}
\maketitle
\def\thefootnote{*}\footnotetext{Equal contribution}\def\thefootnote{\arabic{footnote}}
\begin{abstract}
Information Retrieval (IR) methods aim to identify documents relevant to a query, which have been widely applied in various natural language tasks. However, existing approaches typically consider only the textual content within documents, overlooking the fact that documents can contain multiple modalities, including images and tables. Also, they often segment each long document into multiple discrete passages for embedding, which prevents them from capturing the overall document context and interactions between paragraphs. To address these two challenges, we propose a method that holistically embeds documents interleaved with multiple modalities by leveraging the capability of recent vision-language models that enable the processing and integration of text, images, and tables into a unified format and representation. Moreover, to mitigate the information loss from segmenting documents into passages, instead of representing and retrieving passages individually, we further merge the representations of segmented passages into one single document representation, while we additionally introduce a reranking strategy to decouple and identify the relevant passage within the document if necessary. Then, through extensive experiments on diverse IR scenarios considering both the textual and multimodal queries, we show that our approach substantially outperforms relevant baselines, thanks to the consideration of the multimodal information within documents.
\end{abstract}
\section{Introduction}

Information Retrieval (IR) is the task of fetching relevant documents from a large corpus in response to a query, which plays a critical role in various real-world applications including web search engines and question-answering systems~\citep{Shah2019kvqa,Lewis2020rag,Guu2020realm}. Over the years, IR methods have evolved significantly, broadly categorized into sparse and dense retrieval paradigms. Specifically, sparse retrieval methods~\cite{Robertson1994bm25, Jones2004tfidf} focus on lexical overlap between queries and documents; meanwhile, dense retrieval methods~\cite{Karpukhin2020dpr, Xiong2021ance} utilize neural embeddings to represent queries and documents in a continuous vector space. Note that, recently, dense retrieval methods have gained more popularity over sparse methods due to their capability to capture semantic nuances and context beyond simple keyword matching.

\begin{figure*}[t]
    \centering
    \includegraphics[width=\linewidth]{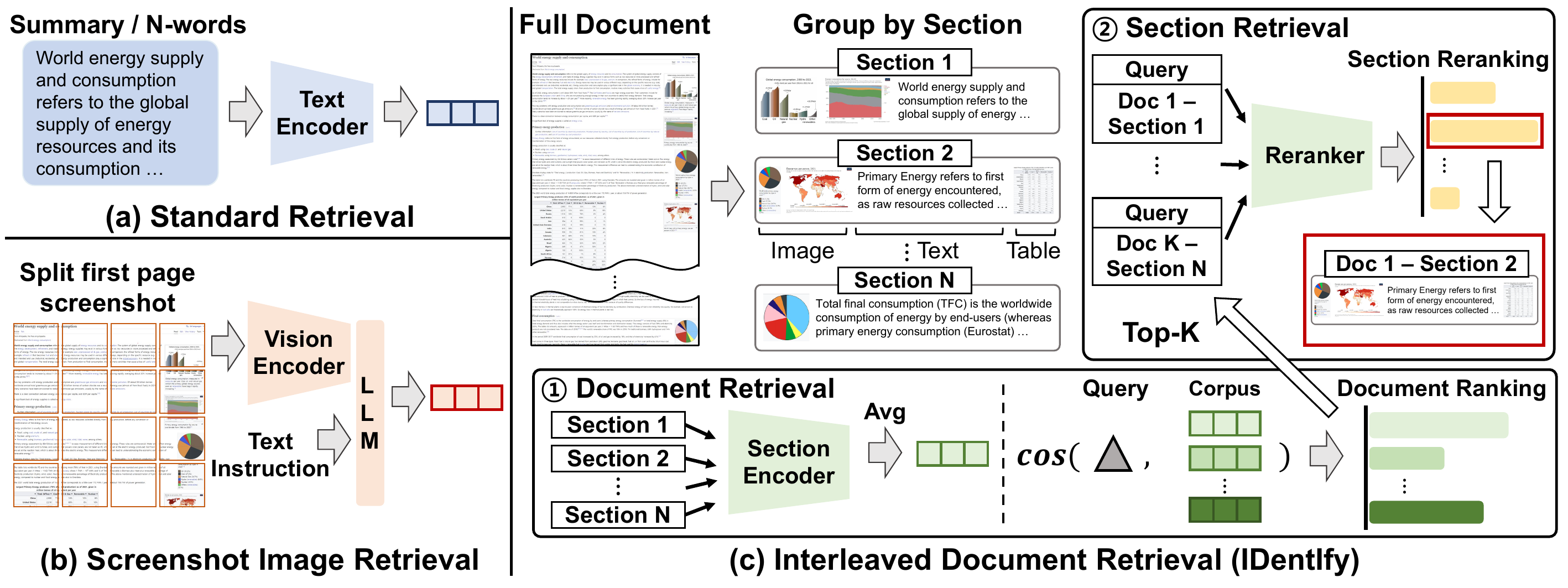}
    \par
    \caption{Comparison of different IR approaches. \textbf{(a)}: Conventional methods use a small portion of the text within the document for its representation. \textbf{(b)}: Recent methods use first-page screenshot images to represent the document. \textbf{(c)}: Our approach leverages the full contextual information within documents interleaved with multiple modalities by considering them in their original format, and is further capable of pinpointing relevant sections for the query.}
    \label{fig:IR_comparison}
\end{figure*}

Despite their successes, existing (dense) retrieval methods face a couple of severe challenges. First, they primarily rely on the textual data for document embedding and retrieval, overlooking the fact that modern documents often contain multimodal content, including images and tables (beyond the plain text), which can carry information that may be essential for accurately understanding and retrieving the relevant documents~\cite{multimodalarxiv}. For instance, a diagram within a medical article can more effectively represent the structure of a molecule or the progression of a disease, offering more clarity that would be difficult to achieve with text alone, and omitting such multimodal content can lead to an incomplete understanding (and potentially inaccurate retrieval) of the documents. Also, the segmentation of long documents into discrete passages, which is commonly employed by existing retrieval models to handle the length limitation for embeddings~\citep{Karpukhin2020dpr, Xiong2021ance}, may prevent models from capturing the full context and the intricate relationships between different parts of the document, ultimately leading to suboptimal retrieval performance~\citep{DBLP:conf/emnlp/DongDLZLZL24, LongRAG}. Notably, concurrent to our work, while there has been recent work that screen captures the document and then embed its screenshots (to consider different modalities in a unified format)~\citep{Faysse2024colpali,Ma2024dse}, not only its content (such as paragraphs, images, and tables) can be fragmented into different sub-images, leading to the loss of contextual coherence across the entire document, but also the visual representation of text may hinder the model's ability to capture the semantic relationships present in the original textual data, and increasing image resolution raises concerns on memory requirements.

To tackle these challenges, we introduce a novel approach to holistically represent documents for IR, representing and retrieving documents interleaved with multiple modalities in a unified manner (illustrated in Figure~\ref{fig:IR_comparison}). Specifically, it revolves around the recent advance of Vision-Language Models (VLMs), which enable the processing and integration of multimodal content (such as text, images, and tables) directly into a single token sequence, thereby preserving the context and relationships between various parts of the document, unlike prior methods that rely on the fragmented visual representations. Additionally, in cases where the number of tokens in a document is large and exceeds the capacity of a single context window of VLMs, we propose a strategy to segment the document into passages, each represented within the token limit, and combine these passage embeddings into a unified document representation. This strategy differs from existing approaches that independently represent and retrieve at the passage level, potentially losing the overall document context. Lastly, to accurately identify only the relevant sections within the retrieved lengthy document, we introduce a reranking mechanism that is trained to pinpoint the passage most pertinent to the query (among all the other passages within the document), allowing for both the coarse-grained document-level matching and fine-grained passage-level retrieval. We refer to our overall framework as \textbf{I}nterleaved \textbf{D}ocum\textbf{ent} \textbf{I}n\textbf{f}ormation Retrieval S\textbf{y}stem (\ours).

We experimentally validate the effectiveness of \ours on four benchmark datasets, considering both text-only and multimodal queries. On a battery of tests conducted, we observe that our approach substantially outperforms relevant baselines that consider only the uni-modality or certain facets of multi-modality, thanks to the holistic consideration of interleaved multimodal contents. Furthermore, we find that the strategy to represent the whole document with its single representation (by merging embeddings of its splits) is superior to the approach of individually representing them for document retrieval, but also performing reranking over the sections of the retrieved document is superior to the approach of directly retrieving those sections, confirming the efficacy of proposed coarse-to-fine retrieval and reranking pipeline for document and passage retrieval, respectively.

\section{Related Work}
\label{sec:related_work}

\paragraph{Information Retrieval} 
IR involves finding documents relevant to a query, which plays a crucial role in applications such as search and question-answering~\citep{ir_survey, ir_survey2, DBLP:journals/tacl/RamLDMSLS23, DBLP:conf/naacl/ShiMYS0LZY24, DBLP:conf/naacl/JeongBCHP24}. Earlier IR approaches measured the similarity between queries and documents based on their lexical term matching, such as BM25 and TF-IDF~\citep{Robertson1994bm25, Jones2004tfidf}. However, these methods struggled to capture semantic nuances beyond surface-level term overlaps. Recently, along with advancements in language models~\citep{bert, roberta}, there have been dense retrieval methods that embed both queries and documents into a shared dense vector space~\citep{Karpukhin2020dpr, Xiong2021ance}, enabling the calculation of semantic similarity between them more effectively by capturing the deeper contextual information. Yet, previous studies have mainly focused on enhancing the textual representations of queries and documents, while overlooking the multimodal nature of documents beyond text, which can provide richer context and aid in more accurate retrieval~\citep{DBLP:conf/iccv/0002OTG21, DaQu2024Jeong}.

\paragraph{Multimodal Information Retrieval} 
Recent studies in IR have expanded the focus from purely text-based retrieval models to those that consider other modalities, such as images~\citep{Radford2021clip, Xiao2024autover}, tables~\citep{HerzigMKE21, ChenZR24} and graphs~\citep{Baek2023difar}; however, the majority of these approaches~\citep{Zhou2024vista,Long2024genmulti, Lerner2024multimodalretrieval, DBLP:conf/naacl/NowakPA24, Caffagni2024wikillava} have primarily explored how to process the multimodal \textit{queries}, and overlooked the equally important multimodal characteristics of the \textit{documents} being retrieved. In efforts to handle diverse multimodal elements within documents, there are concurrent studies that have proposed to capture screenshots of documents, such as PDFs~\citep{Faysse2024colpali, M3DocRAG} or Wikipedia web pages~\citep{Ma2024dse}, and subsequently encoding them through vision models~\citep{Ding2024pdfmvqa}. Yet, these methods are not only limited by factors, such as image resolution and computational memory, constraining their application to documents longer than a single page\footnote{It requires processing 9.8k image tokens just to process a single-page document, and it results in 2TB of storage for handling the entire Wikipedia corpus, which may not be practical.}, but also fall short by treating the diverse modalities within a document as a single visual entity, leading to document representations that may fail to effectively capture the nuanced interdependence between text and images. Also, while there are concurrent studies~\cite{VLM2Vec, MM-Embed} that consider images and text as retrieval targets, they primarily focus on representing image-text pairs and their retrieval, rather than addressing the holistic representation of documents that include multiple images and another modality (tables). Finally, all the aforementioned work does not address the issue of splitting documents into smaller fragments (passages or sub-images), which may disrupt the holistic contextual view of the entire document. 

\paragraph{Vision-Language Models}
Recently developed VLMs have emerged as a powerful tool for jointly processing visual and textual data, which combine the image understanding capabilities of visual encoders~\citep{Radford2021clip,Zhai2023siglip} with the advanced reasoning abilities of language models~\citep{ChatGPT,gpt4}. They have achieved remarkable performance across diverse vision-language tasks (such as image captioning and visual question answering)~\citep{Dai2023instructblip, gpt4v}, with substantially limited attention on their applications to IR. We note that the latest developments in this field have particularly focused on enabling VLMs to handle interleaved, multimodal content, involving a mixed sequence of images and text~\citep{Zhang2023internlmXcomposer, Li2024llavanextinterleave}. In particular, LLaVA-NeXT-Interleave~\citep{Li2024llavanextinterleave} introduces a fine-tuning approach that specifically enhances the VLMs' capacity to understand complex interleavings of multiple images and text in a single context. Drawing inspiration from these advances, we propose to harness their capabilities to create unified embeddings for documents interleaved with text and images (and tables).

\section{Method}\label{headings}
We present \ours to holistically represent documents interleaved with multimodal elements. 

\subsection{Preliminaries}\label{sec:sub:prelim}
We begin with formally explaining IR and VLMs.

\paragraph{Information Retrieval}
Information Retrieval (IR) is the task of identifying a set of relevant documents $\{\vd_1, \vd_2, \ldots, \vd_k\} \subseteq \mathcal{D}$ from a large corpus $\mathcal{D}$, given a query $\vq$. Here, each query $\vq$ and document $\vd$ are represented as a sequence of tokens, e.g., $\vq = [q_1, \ldots, q_n]$, and traditional IR approaches typically consider these tokens as purely textual elements. However, we propose to extend this assumption to include the tokens of both the textual and visual content, to capture the multimodal nature of many real-world documents. Then, this new extension raises important questions of how can both the textual and visual content be represented within a unified token framework, and how can these multimodal tokens be seamlessly integrated and encoded for document representations.

\begin{figure*}[t]
    \centering
    \includegraphics[width=0.875\linewidth]{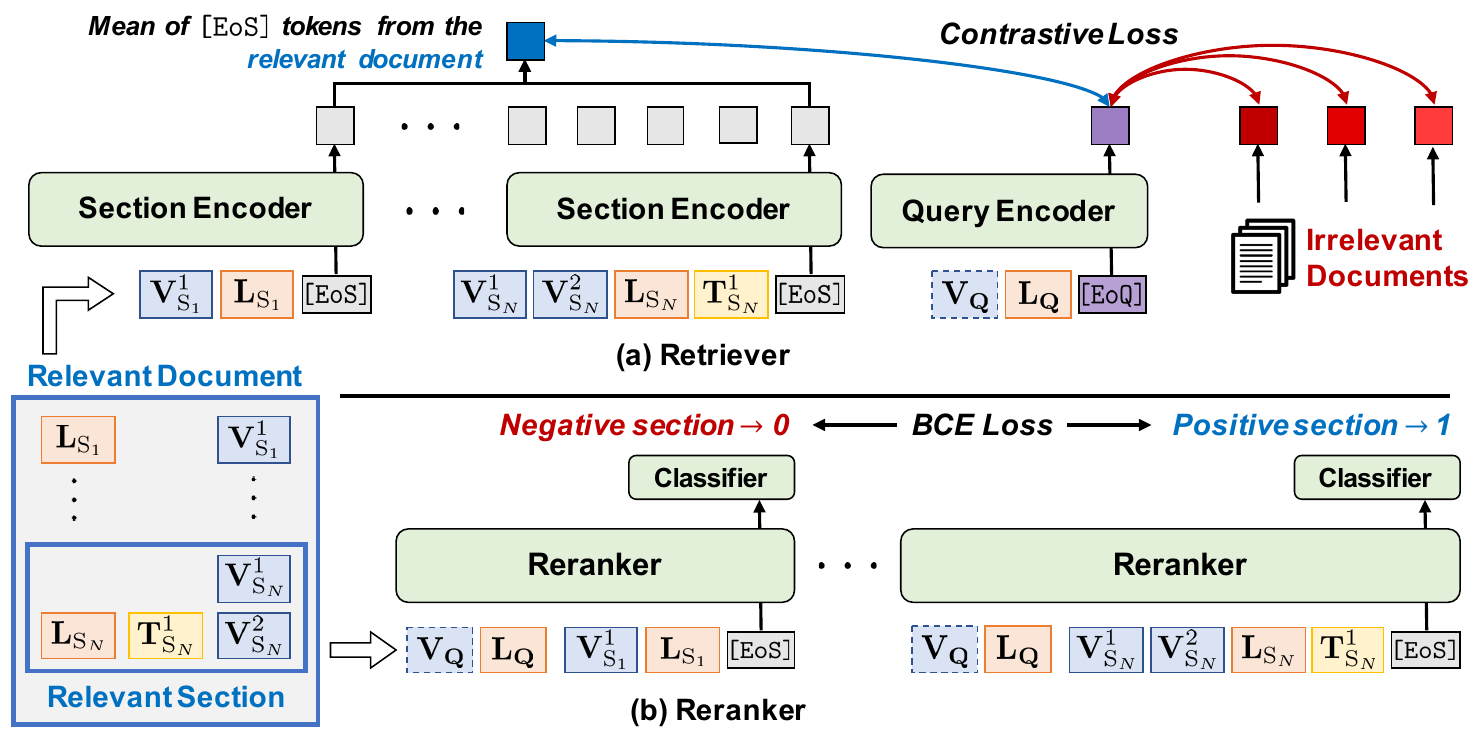}
    \par
    \caption{\small Overview of the proposed \ours. \textbf{(a)}: In our document retriever, a query encoder represents a query (\textcolor{violet}{purple}), and sections are encoded with a section encoder whose embeddings are averaged to form a document representation (\textcolor{blue}{blue}). Contrastive learning loss (\textcolor{red}{red}) is used for training the document retriever. \textbf{(b)}: Reranker scores query-section relevance with the concatenation of the query and section, trained using Binary Cross-Entropy loss.}
    \label{fig:method}
\end{figure*}

\paragraph{Vision-Language Models}
To answer them, we now turn to describing VLMs, which are designed to jointly encode the textual and visual information in a unified token framework. These models are generally comprised of two main components: a visual encoder and a language model, interconnected through a projection layer. Specifically, given the document that may contain interleaved modalities (e.g., text and images), the visual encoder extracts high-level visual features from images embedded within the document, mapping them into a latent space. Then, these visual features are transformed into a sequence of visual tokens via the projection layer, represented as follows: $\mathbf{V}\!\in\!\mathbb{R}^{V\!\times\!d_{\text{emb}}}$, where $V$ denotes the visual token length and $d_{\text{emb}}$ is the token dimension size. Similarly, for the textual content embedded within the document, the language model uses a word embedding layer to convert the input text into a sequence of tokens, as follows: $\mathbf{L}\!\in\!\mathbb{R}^{L\!\times\!d_{\text{emb}}}$, where $L$ denotes the text token length.

In this work, we also propose to account for tables that are the integral modality to holistically represent the full content of documents. Yet, unlike text and images that have dedicated processing layers within VLM architectures, tables do not have a specific representation layer. Nevertheless, we argue that VLMs are pre-trained on diverse web data, and subsequently learned implicitly to handle the table structures formatted in HTML. Consequently, we treat HTML-format table data as a linearized sequence of HTML words, applying the same word embedding layer as is used for plain text. To be formal, this process converts the table content into table tokens, as follows: $\mathbf{T}\!\in\mathbb{R}^{T\!\times\!d_{\text{emb}}}$, where $T$ is the token length of the table. Lastly, once extracted, the visual tokens, text tokens, and table tokens are concatenated (into a unified token sequence) and then passed through the remaining layers of VLMs, to capture both uni- and cross-modal relationships across different modalities, ultimately enabling the comprehensive understanding of the documents.

\subsection{Retriever}\label{sec:sub:retriever}
We now explain how we design a retriever specifically tailored for multimodal interleaved document retrieval. In particular, our approach leverages a VLM capable of processing text, images, and tables within a single document. Further, following the standard practice of existing retrieval architectures~\citep{Karpukhin2020dpr, Xiong2021ance}, we use a dual-encoder structure, which consists of a query encoder and document (or section) encoder, both are based on VLMs, illustrated in~\Cref{fig:method} \highlight{(a)}.

Specifically, thanks to the use of the VLM, our query encoder can take either purely textual queries $\vq\!=\!\mathbf{L}_{\mathrm{Q}}$ or multimodal queries consisting of text and visual elements $\vq\!=\![\mathbf{V}_{\mathrm{Q}},\;\mathbf{L}_{\mathrm{Q}}]$. Also, to obtain the final query representation, we use a learnable token called `End of Query', $\texttt{[EoQ]}\!\!\in\!\mathbb{R}^{d_{\text{emb}}}$, which is appended to the end of the query tokens $\vq$. The final concatenated tokens $\left[\vq,\;\texttt{[EoQ]}\right]$ are then passed through the query encoder. Lastly, the model output corresponding to $\texttt{[EoQ]}$ is used as the final query representation, as follows: $\mathbf{Z}_\mathrm{Q}\!\in\!\mathbb{R}^{d_\text{emb}}$.

For documents, we represent each of them $\vd$ as a sequence of sections: $\vd\!=\![\vs_i]_{i=1}^{S}$ (with a total of $S$ sections), where each section $\vs_i$ is derived by dividing the document according to its subtitles. $\vs_i$ can contain a combination of text tokens $\mathbf{L}_{\mathrm{S}i}$, visual tokens from embedded images $\mathbf{V}_{\mathrm{S}i}$, and table tokens $\mathbf{T}_{\mathrm{S}i}$, denoted as follows: $\vs_i\!=\!\left[\mathbf{V}_{\mathrm{S}_i},\;\mathbf{L}_{\mathrm{S}_i},\;\mathbf{T}_{\mathrm{S}_i}\right]$. Then, to obtain a section-level representation, similar to the query representation, we introduce a learnable token, called `End of Section': $\texttt{[EoS]}\!\!\in\!\mathbb{R}^{d_{\text{emb}}}$, which is appended at the end of each section. We then forward concatenated tokens $\left[\vs_i,\;\texttt{[EoS]}\right]$ to the section encoder, and, after that, the output corresponding to \texttt{[EoS]} is used to form the section representation, as follows: $\mathbf{Z}_{\mathrm{S}_{i}}\!\in\!\mathbb{R}^{d_{\text{emb}}}$. Additionally, the overall document representation is obtained by averaging the representations of all sections within the document, as follows: $\mathbf{Z}_{\mathrm{D}}\!=\!\frac{1}{S}\!\sum_{i=1}^{S}\mathbf{Z}_{\mathrm{S}_i}$.

The remaining step is to train those two query and section encoders. Recall that the goal of the retriever is to assess a relevance score between the query and the document. To achieve this, we use a contrastive learning loss based upon the query and document representations, whose objective is to assign higher similarity scores to relevant documents (positive samples) and lower scores to irrelevant ones (negative samples) for the query, as follows:
\begin{align}
    \mathcal{L}_\text{retriever}\!&=\!-\frac{1}{B}\sum^{B}_{i=1}\log\left(\frac{\phi\left(\mathbf{Z}_{\mathrm{Q}_{i}},\mathbf{Z}_{\mathrm{D}_{i}}\right)}{\sum^{B}_{j=1}\phi\left(\mathbf{Z}_{\mathrm{Q}_{i}}, \mathbf{Z}_{\mathrm{D}_{j}}\right)}\right), \notag \\\phi\left(\mathbf{a},\;\mathbf{b}\right)\!&=\!\exp\left(\frac{\mathbf{a}^{\top}\mathbf{b}}{\lVert \mathbf{a} \rVert\lVert \mathbf{b} \rVert}\right),
    \label{eq_contrast}
\end{align}
where $B$ is the batch size. By minimizing $\mathcal{L}_\text{retriever}$, the retriever learns to optimize the similarity between queries and their relevant documents, enabling the retrieval of the most pertinent documents for the given input query during inference.

\subsection{Reranker}\label{sec:sub:reranker}
To enable fine-grained retrieval within documents beyond the retrieval of documents themselves, we introduce a section-level reranking mechanism that identifies the section most relevant to the query. In particular, once the document is retrieved, the objective of the reranker $f_{\mathrm{R}}$ is to pinpoint the specific sections within the document that best match the query. We also note that this reranker is similarly operationalized with the VLM along with a binary classifier on top of it, which directly measures the relevance of each query-section pair (\Cref{fig:method} \highlight{(b)}).

Formally, for a retrieved document, we take each of its sections $\vs_i$ with a learnable token for section embedding \texttt{[EoS]} attached to the end and concatenate it with query $\vq$ , forming the input sequence of $\left[\vq,\;{\vs}_i,\;\texttt{[EoS]}\right]$. The concatenated tokens are then processed through the reranker, and its output corresponding to \texttt{[EoS]} captures the relevance between the query and section, which is further subsequently passed to a binary classifier. Through this, the classifier outputs a probability score indicating the likelihood of the section being relevant to the query, \textit{i.e.}, a score close to one denotes a high relevance.

To train this reranker, we use the binary cross-entropy loss, formulated as follow:
\begin{align}
    \mathcal{L}_\text{reranker}&\!=\!\sum_{i=1}^{B}\sum_{j=1}^{S_{i}}\!\frac{1}{BS_{i}}\ell\left(\mathbf{y}_{\vs_{i,j}},\;f_{\mathrm{R}} \left(\left[\vq,\;\hat{\vs}_{i,j} \right] \right)\right), \notag \\
    \ell\left(y,\hat{y}\right)&\!=\!-\left[y\log\hat{y}\!+\!(1\!-\!y)\log(1\!-\!\hat{y})\right],
    \label{eq_reranker_bce}
\end{align}
where $S_{i}$ is the number of sections in the $i$-th document, $\mathbf{y}_{\vs_{i,j}}$ is the label for the $j$-th section of the $i$-th document $\vs_{i, j}$ (with its value of one if relevant to the query $\vq$, otherwise zero), $\hat{\vs}_{i,j}=\left[\vs_{i,j},\;\texttt{[EoS]}\right]$, and $B$ is the batch size during training. Also, during training, the sections not labeled as relevant to the query are considered negative samples. Then, by minimizing $\mathcal{L}_{\text{reranker}}$, the reranker learns to predict section relevance for any query, thereby refining our retrieval process by allowing the retrieval of not just whole documents but also their most relevant sections, for multiple use cases of IR.

\section{Experiments}\label{sec:experiments}

\subsection{Experimental Setups}\label{sec:experiments:datasets}

\paragraph{Datasets}
We evaluate the proposed \ours on four benchmark datasets designed for multimodal IR that require understanding of both textual and visual cues within queries and documents, as follows: \textbf{Encyclopedic-VQA}~\citep{Mensink2023encyc} is a large-scale benchmark for multimodal Visual Question Answering (VQA) with queries linked to specific Wikipedia sections and includes both textual and multimodal queries; \textbf{InfoSeek}~\citep{chen2023infoseek} is a knowledge-intensive VQA dataset with multimodal questions generated from Wikidata triples that include diverse entities such as landmarks, animals, and food; \textbf{ViQuAE}~\citep{Lerner2022viquae} involves both textual and multimodal queries about human entities, linked to annotated Wikipedia sections, making it ideal for evaluating section reranking; \textbf{Open-WikiTable}~\citep{Kweon2023openwikitable} targets open-domain table QA by identifying documents or sections containing relevant tables. We provide more details on datasets in~\Cref{appendix:Implementation Details}.

\paragraph{Baselines}

To comprehensively validate \ours, we compare it against two categories of baselines:
\begin{itemize}[itemsep=0.5mm, parsep=1pt, leftmargin=*]
\item \textbf{Conventional VLM Baselines:} We consider earlier VLMs, which are not capable of jointly processing text and images, such as CLIP~\cite{Radford2021clip} and BLIP~\citep{Li2023blip}. Also, we consider the approaches, such as UniIR~\citep{Wei2023unir}, which is built on top of them and fine-tuned with a contrastive loss (\Cref{eq_contrast}). These baselines serve as reference points to assess performance gains from recent VLM advances rather than serving as direct competitors.

\item \textbf{Baselines with Different Document Representations:} We further consider existing approaches, representing documents in various ways. \textbf{Entity} and \textbf{Abstract} baselines retrieve documents based on their titles and summaries, respectively, using high-level textual cues. \textbf{Text-only} baselines utilize the full textual content of documents for retrieval~\citep{Caffagni2024wikillava,Wang2024dapr}. \textbf{Text \& Table} and \textbf{Text \& Image} baselines leverage tables and first image of documents alongside the text, respectively~\citep{Jiang2024e5v,MM-Embed,VLM2Vec}. \textbf{\ours} is our model that holistically represents multimodal content (text, images, and tables) in documents. All baselines share the same recent VLMs as \ours, allowing for a controlled comparison focused on document representation strategies.
\end{itemize}

\begin{table}[t]
\centering
\resizebox{0.475\textwidth}{!}{
    \renewcommand{\arraystretch}{1.0}
    \renewcommand{\tabcolsep}{2.0mm}
    \begin{tabular}{l c c c c}
         \toprule
         {\textbf{Method}} & {\textbf{R@1}} & {\textbf{R@10}} & {\textbf{R@100}} & {\textbf{MRR@10}}\\
         \midrule
         \multicolumn{5}{c}{\cellcolor{gray}{CLIP-VIT-L-14}}\\
         Zero-Shot & 1.9 & 6.3 & 13.9 & 3.1 \\
         UniIR + Text-Only & 3.8 & 20.6 & 50.3 & 7.7 \\
         UniIR + Text \& Image & 5.8 & 21.5 & 48.5 & 10.0 \\
         \multicolumn{5}{c}{\cellcolor{gray}{BLIP-Large}}\\
         Zero-Shot & 0.0 & 0.0 & 0.0 & 0.0 \\
         UniIR + Text-Only & 9.8 & 36.9 & 71.4 & 16.3 \\
         UniIR + Text \& Image & 9.9 & 23.9 & 60.7 & 13.5 \\
        \multicolumn{5}{c}{\cellcolor{gray}{LLaVA-NeXT-Interleave-0.5B}}\\
         Entity &
         {3.1} & {15.5} & {39.7} & {6.1}\\
         Abstract &
         {13.4} & {41.3} & {66.5} & {21.6}\\
         Text-Only &
         {12.5} & {37.8} & {68.7} & {19.8}\\
         Text \& Table &
         {12.6} & {38.6} & {68.5} & {19.9}\\
         Text \& Image &
         {16.4} & {45.4} & {77.1} & {25.3}\\
         \cellcolor{gg}\ours (Ours) &
         \cellcolor{gg}{\textbf{20.5}} & \cellcolor{gg}{\textbf{50.0}} & \cellcolor{gg}{\textbf{78.0}} & \cellcolor{gg}{\textbf{29.4}}\\
         \bottomrule
    \end{tabular}
}
\caption{Results with different document retrievers.}
\label{tab:doc_format}
\end{table}

\paragraph{Evaluation Metrics}
To evaluate our approach, we use standard metrics: Recall@K (R@K) measures whether the relevant document or section appears within the top-K results; MRR@K measures how early the first relevant item is ranked (within top-K) by averaging its inverse rank across queries. 

\paragraph{Implementation Details}
We use LLaVA-NeXT-Interleave~\citep{Li2024llavanextinterleave} as the basis VLM for both the retriever and reranker, and also use LLaVA-OneVision~\citep{Li2024llavaov} as an additional basis VLM to show the robustness of \ours. Following the convention of using the basis of retrieval with less than 1B parameters to balance computational efficiency and retrieval performance~\citep{Radford2021clip,Zhou2024vista,Wei2023unir}, we choose 0.5B-parameter versions of the VLMs. During training, documents are represented using randomly selected four sections, while in inference, we consider all sections within each document. For section-level retrieval, all sections within the top 25 retrieved documents are reranked. Experiments are conducted on a single H100 GPU.

\subsection{Experimental Results and Analyses}\label{sec:experiments:exp_results}

\begin{table}[t]
\centering
\resizebox{0.475\textwidth}{!}{
\renewcommand{\arraystretch}{0.9}
\renewcommand{\tabcolsep}{2.5mm}
\begin{tabular}{l c c c c}
     \toprule
     {\textbf{Granularity}} & {\textbf{R@1}} & {\textbf{R@10}} & {\textbf{R@20}} & {\textbf{MRR@10}}\\
     \midrule
     Passage* &
     {3.9} & {16.9} & {22.0} & {7.5}\\
     Passage &
     {28.6} & {36.4} & {37.8} & {31.2}\\
     \cellcolor{gg}Document (Ours) &
     \cellcolor{gg}{\textbf{35.1}} & \cellcolor{gg}{\textbf{50.8}} & \cellcolor{gg}{\textbf{53.6}} & \cellcolor{gg}{\textbf{40.3}}\\
     \bottomrule
\end{tabular}
}
\caption{Comparison of different IR strategies for section retrieval. Document (Ours) performs document retrieval and section reranking, whereas Passage performs passage retrieval and reranking. * denotes the model without reranking.}
\label{tab:pas_doc}
\end{table}

\paragraph{Main Results}
We report retrieval performance on the Encyclopedic-VQA dataset in \Cref{tab:doc_format}, where queries include both text and images. \ours significantly outperforms all baselines built on VLMs such as CLIP and BLIP, which are limited to handling a single image alongside text and encoding image-text representations independently, making them suboptimal for understanding multimodal interactions within documents. We also observe that \ours achieves the best performance, improving R@1 scores by 53.0\%, 64.0\%, 62.7\%, and 25.0\% over Abstract, Text-Only, Text \& Table and Text \& Image retrieval baselines, respectively, with similar trends observed for other metrics. These results demonstrate the effectiveness of integrating multimodal content holistically into a unified representation. To further illustrate the advantages of our approach, we provide case studies in~\Cref{appendix:case studies}.

\begin{table*}[t]
\centering
\tiny
\begin{minipage}{\textwidth}
    \centering
    \resizebox{0.99\textwidth}{!}{
    \renewcommand{\arraystretch}{0.75}
    \renewcommand{\tabcolsep}{3.5pt}
    \begin{tabular}{lcl c cccc c cccc}
    \toprule
    & & & & \multicolumn{4}{c}{\textbf{Document Retrieval}} && \multicolumn{4}{c}{\textbf{Section Reranking}} \\
    \textbf{Dataset} & \textbf{Query Type} & \textbf{Method} &\;& \textbf{R@1} & \textbf{R@10} & \textbf{R@100} & \textbf{MRR@10} &\;& \textbf{R@1} & \textbf{R@10} & \textbf{R@20} & \textbf{MRR@10} \\
    \midrule
    {\multirow{4}{*}{Enc-VQA}} & {\multirow{2}{*}{Multimodal}} & {Text-Only} && {12.5} & {37.8} & {68.7} & {19.8} && {40.7} & {52.8} & {55.5} & {44.8} \\
    & & {\ours (Ours)} && \textbf{20.5} & \textbf{50.0} & \textbf{78.0} & \textbf{29.4} && \textbf{42.4} & \textbf{53.6} & \textbf{55.7} & \textbf{46.3} \\
    \cmidrule{2-13}
    & {\multirow{2}{*}{Textual}} & {Text-Only} && {62.7} & {76.3} & {87.4} & {67.0} && {68.1} & {79.4} & {80.2} & {72.3} \\
    & & {\ours (Ours)} && \textbf{65.4} & \textbf{76.8} & \textbf{87.8} & \textbf{69.0} && \textbf{69.7} & \textbf{80.1} & \textbf{80.6} & \textbf{73.6} \\
    \midrule
    {\multirow{4}{*}{ViQuAE}} & {\multirow{2}{*}{Multimodal}} & {Text-Only} && {13.5} & {40.4} & {67.4} & {20.9} && \textbf{12.6} & {31.7} & {37.7} & \textbf{18.2} \\
    & & {\ours (Ours)} && \textbf{17.5} & \textbf{46.0} & \textbf{69.4} & \textbf{26.3} && {11.4} & \textbf{32.1} & \textbf{39.2} & {17.5} \\
    \cmidrule{2-13}
    & {\multirow{2}{*}{Textual}} & {Text-Only} && {55.8} & {71.5} & {83.0} & {60.9} && {27.8} & {50.2} & {57.7} & {35.0} \\
    & & {\ours (Ours)} && \textbf{56.5} & \textbf{72.2} & \textbf{83.0} & \textbf{61.6} && \textbf{29.9} & \textbf{50.9} & \textbf{59.8} & \textbf{36.7} \\
    \midrule
    {\multirow{2}{*}{InfoSeek}} & {\multirow{2}{*}{Multimodal}} & {Text-Only} && {6.8} & {23.6} & {52.5} & {11.2} && {N/A} & {N/A} & {N/A} & {N/A} \\
    & & {\ours (Ours)} && \textbf{10.2} & \textbf{30.4} & \textbf{57.3} & \textbf{15.7} && {N/A} & {N/A} & {N/A} & {N/A} \\
    \bottomrule
    \end{tabular}}
    \caption{\small Performance on document retrieval and section reranking for multimodal and textual queries on Encyclopedic-VQA (Enc-VQA), ViQuAE, and InfoSeek. We compare the approach that solely uses textual information from documents (Text-Only) and our approach of leveraging interleaved multimodal contents from the documents (\ours) over various scenarios.}
    \vspace{0.15in}
    \label{tab:information_retrieval}
\end{minipage}
\par
\begin{minipage}{\textwidth}
    \centering
    \begin{minipage}{0.42\textwidth}
        \begin{minipage}{\linewidth}
            \centering
            \small{\textbf{(a) Document Retrieval for Tables}}\\
            \vspace{0.025in}
            \resizebox{\textwidth}{!}{
                \renewcommand{\arraystretch}{0.8}
                \renewcommand{\tabcolsep}{2.0mm}
                \begin{tabular}{l c c c c}
                     \toprule
                     {\textbf{Method}} & {\textbf{R@1}} & {\textbf{R@10}} & {\textbf{R@100}} & {\textbf{MRR@10}}\\
                     \midrule
                     {Zero-shot} &
                     {29.4} & {58.0} & {86.0} & {38.1}\\
                     {Finetuned} & {\textbf{55.8}} & {\textbf{84.1}} & {\textbf{93.5}} & {\textbf{66.1}}\\
                     \bottomrule
                \end{tabular}
            }
        \end{minipage}
        \par
        \vspace{-0.025in}
        \begin{minipage}{\linewidth}
            \centering
            \vspace{0.075in}
            \small{\textbf{(c) Tabular Classification}}\\
            \vspace{0.025in}
            \resizebox{\textwidth}{!}{
                \renewcommand{\arraystretch}{0.75}
                \renewcommand{\tabcolsep}{3.0mm}
                \begin{tabular}{l c c c}
                     \toprule
                     {\textbf{Method}} & {Random} & {Zero-shot} & {Finetuned}\\
                     \midrule
                     {\textbf{Acc@1}} & {11.9} & {9.3} & {\textbf{56.5}}\\
                     \bottomrule
                \end{tabular}
            }
        \end{minipage}
        
    \end{minipage}
    \begin{minipage}{0.56\textwidth}
        \centering
        \small{\textbf{(b) Section Reranking for Tables}}\\
        \vspace{0.025in}
        \resizebox{\textwidth}{!}{
            \renewcommand{\arraystretch}{1.3}
            \renewcommand{\tabcolsep}{1.5mm}
            \begin{tabular}{l c l c c c c}
                 \toprule
                 {\textbf{Dataset}} &{\textbf{Target}} & {\textbf{Method}} & {\textbf{R@1}} & {\textbf{R@10}} & {\textbf{R@20}} & {\textbf{MRR@10}}\\
                 \midrule
                 {\multirow{2}{*}{ViQuAE}} & {\multirow{2}{*}{Text}} & {Zero-shot} & {20.3} & {49.0} & {57.7} & {28.9}\\
                  & & {Finetuned} & \textbf{29.9} & {\textbf{50.9}} & {\textbf{59.8}} & {\textbf{36.7}}\\
                 \midrule
                 {\multirow{2}{*}{OWT}} & {\multirow{2}{*}{Table}} & {Zero-shot} & {5.9} & {20.5} & {29.4} & {9.1}\\
                  & & {Finetuned} & \textbf{8.4} & {\textbf{36.7}} & {\textbf{52.8}} & {\textbf{15.2}}\\
                 \bottomrule
            \end{tabular}
        }
    \end{minipage}
    \caption{Retrieval results for tables, where Zero-shot denotes a model trained on Encyclopedic-VQA but not on the target dataset. Finetuned refers to additional training of the model on the target dataset. \textbf{(a)}: Results for tabular document retrieval on Open-WikiTable (OWT). \textbf{(b)}: Textual and tablular section reranking results on ViQuAE and OWT datasets, respectively. \textbf{(c)}: Reranker accuracy of a classification task that identifies the section containing the query-associated table given a gold document.}
    \label{tab:tab_info_ret}
\end{minipage}
\end{table*}

We further examine the impact of our pipeline of document retrieval and section reranking. In \Cref{tab:pas_doc}, the passage retriever represents individual sections as separate retrieval units, whereas the document retriever (ours) aggregates multiple section representations into a single representation. Then, we perform reranking over the retrieved sections or the sections from the retrieved documents, and then report the results in \Cref{tab:pas_doc} (where * denotes the model without reranking). From this, we observe that the passage retriever without reranking (Passage*) achieves suboptimal retrieval performance, highlighting the challenge in pinpointing the most relevant section within a document using traditional retrieval methods. In contrast, when the reranker is used alongside the document retriever, the performance significantly surpasses the passage retrieval, demonstrating the effectiveness of our coarse-to-fine document-to-section retrieval strategy. 

\paragraph{Interleaved format enhances document retrieval across modalities.}
We further expand our experiments to two additional datasets, InfoSeek and ViQuAE, and report document retrieval results. As shown in~\Cref{tab:information_retrieval} \highlight{Left}, our model consistently outperforms the Text-document baseline for both the multimodal and textual queries. We attribute these gains to the integration of multimodal content, allowing the VLM to capture richer alignments and leverage pre-existing knowledge for more effective document representation~\citep{Xu2024rir}.

\paragraph{Interleaved format is also beneficial in section retrieval.}
Similarly, we evaluate section retrieval performance on Encyclopedic-VQA and ViQuAE datasets, for both multimodal and textual queries. As shown in~\Cref{tab:information_retrieval} \highlight{Right}, our model outperforms the Text-document baseline in most cases. However, the performance gains over the baseline are smaller compared to the document retrieval setup. This is likely because section reranking focuses on evaluating the relationship between a single section and a query (rather than leveraging the holistic context of the entire document), and individual sections may lack the diverse multimodal information necessary for fully capturing the intent of queries. 

\paragraph{Retrieving tables interleaved within documents is challenging.}
\begin{figure*}[t]
\centering
\tiny
\begin{minipage}[t]{0.32\linewidth}
    \captionsetup{type=figure}
        \centering
        \includegraphics[width=\linewidth]{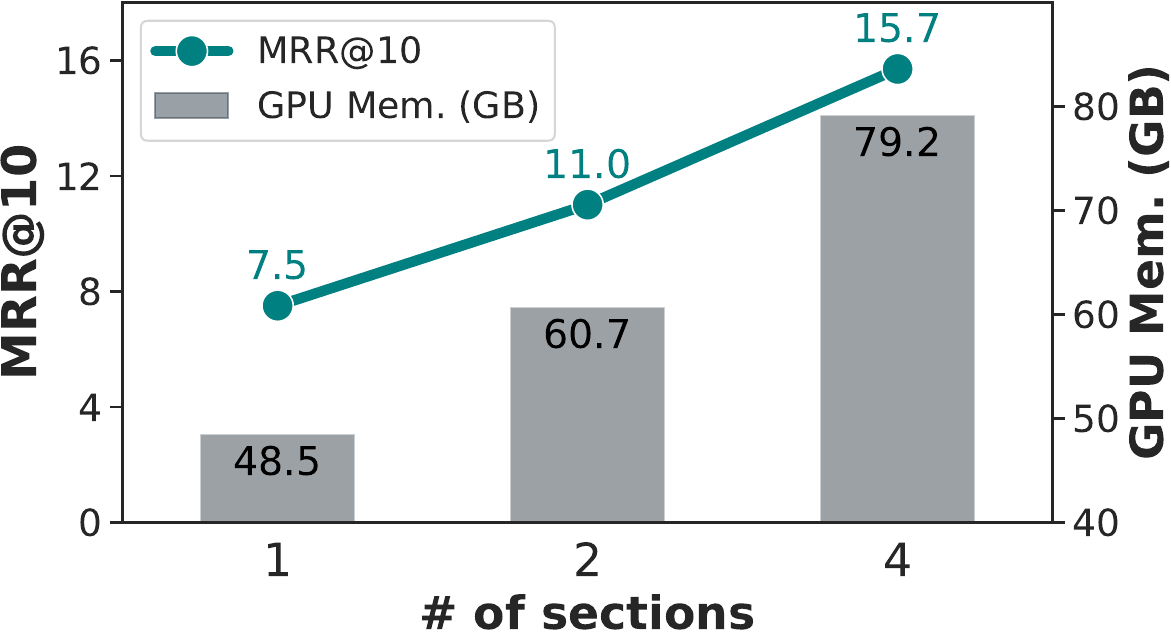}
        \captionof{figure}{\small Trade-off between performance (MRR@10) and training cost (GPU Memory) for retrieval.}
        \label{fig:num_sec}
\end{minipage}
\hspace{0.03in}
\begin{minipage}[t]{0.66\linewidth}
    \captionsetup{type=table}
    \centering
    \resizebox{\textwidth}{!}{
        \renewcommand{\arraystretch}{0.6}
        \renewcommand{\tabcolsep}{2.5mm}
        \begin{tabular}{c l c c c c}
             \toprule
             {\textbf{Quesry Type}} & {\textbf{Train Loss}} & {\textbf{R@1}} & {\textbf{R@10}} & {\textbf{R@20}} & {\textbf{MRR@10}}\\
             \midrule
             {\multirow{3}{*}{Multimodal}} & {Contrastive} & {3.6} & {15.0} & {21.3} & {6.5}\\
             & Doc + BCE & {13.6} & {29.6} & {32.9} & {24.1}\\
             & Sec + BCE (Ours) & {\textbf{42.4}} & {\textbf{53.6}} & {\textbf{55.7}} & {\textbf{46.3}}\\
             \midrule
              {\multirow{3}{*}{Textual}} & {Contrastive} & {13.6} & {37.7} & {45.1} & {20.6}\\
             & Doc + BCE & {23.8} & {43.4} & {47.2} & {39.1}\\
              &   Sec + BCE (Ours) &   {\textbf{69.7}} &   {\textbf{80.1}} &   {\textbf{80.6}} &   {\textbf{73.6}}\\
             \bottomrule
        \end{tabular}
    }
    \captionof{table}{\small Comparison of training objectives for the reranker: Contrastive uses contrastive loss similar to the document retriever training; Doc + BCE concatenates the query with multiple sections from the same document and uses the BCE loss; Sec + BCE trains the reranker by concatenating the query with each section individually.}
\label{tab:rerank_design}
\end{minipage}
\vspace{0.025in}
\end{figure*}

\begin{table*}[t]
\tiny
\centering
\begin{minipage}[t]{0.45\linewidth}
    \vspace{0in}
    \centering
    \resizebox{\linewidth}{!}{
    \renewcommand{\arraystretch}{0.95}
    \renewcommand{\tabcolsep}{2.0mm}
    \begin{tabular}{l c c c}
         \toprule
         {\textbf{Negative}} & {\textbf{R@1}} & {\textbf{R@20}} & {\textbf{MRR@10}}\\
         \midrule
         Top-K &
         {38.1} & {55.3} & {44.4}\\
         In-batch &
         {39.5} & {55.4} & {45.0}\\
         In-document (Ours) &
         {\textbf{42.4}} & {\textbf{55.7}} & {\textbf{46.3}}\\
         \bottomrule
    \end{tabular}
    }
    \caption{\small Comparison of negative sample selection strategies for reranker training: Top-K (top-k retrieved sections), In-batch (sections from other samples in the batch), and In-document (sections in the same document).}
    \label{tab:hard_neg}
\end{minipage}
\hspace{0.04in}
\begin{minipage}[t]{0.53\linewidth}
    \vspace{0in}
    \centering
    \resizebox{\linewidth}{!}{
    \renewcommand{\arraystretch}{0.9}
    \renewcommand{\tabcolsep}{2.5mm}
        \begin{tabular}{l c c c c}
             \toprule
             {\textbf{Format}} & {\textbf{R@1}} & {\textbf{R@10}} & {\textbf{R@100}} & {\textbf{MRR@10}}\\
             \midrule
             Entity &
             {2.3} & {10.3} & {29.7} & {4.3}\\
             Abstract &
             {7.6} & {24.7} & {55.7} & {12.0}\\
             Text-Only &
             {7.0} & {24.1} & {50.4} & {11.7}\\
             Text \& Table &
             {6.9} & {26.3} & {54.9} & {12.1}\\
             Text \& Image &
             {9.3} & {31.4} & {61.9} & {15.4}\\
              \cellcolor{gg}\ours (Ours) &
              \cellcolor{gg}{\textbf{12.1}} & \cellcolor{gg}{\textbf{36.1}} & \cellcolor{gg}{\textbf{62.5}} &  \cellcolor{gg}{\textbf{18.2}}\\
             \bottomrule
        \end{tabular}
    }
    \caption{\small Results with another base model (LLaVA-OneVision-0.5B) for document retrieval (with different document formats).}
\label{tab:llava_ov_doc_format}
\end{minipage}
\end{table*}
We explore the retrieval task for tabular data, aiming to identify documents or sections containing query-relevant tables, and compare models trained on Encyclopedic-VQA (Zero-shot) with those additionally trained on Open-WikiTable (Finetuned). As shown in~\Cref{tab:tab_info_ret} \highlight{(a)}, the Finetuned retriever outperforms the Zero-shot retriever on retrieving documents containing query-relevant tables. However, more fine-grained section reranking results (identifying sections containing query-relevant tables) in~\Cref{tab:tab_info_ret} \highlight{(b)} may reveal a notable modality-specific challenge: the performance of Zero-shot and Finetuned rerankers is considerably lower on table retrieval compared to their performance on text retrieval, despite both the text and tables being represented with word tokens. To better understand this, we design a classification task, where rerankers are tasked with identifying the correct section containing the target table within the golden document. Then, as shown in~\Cref{tab:tab_info_ret} \highlight{(c)}, the Zero-shot reranker performs comparably to random selection, while the Finetuned reranker shows modest improvements. These findings highlight the intrinsic challenge of tabular retrieval, suggesting the need for table-specific modules to more holistically represent multimodal interleaved documents. 

\paragraph{More sections enhance document retrieval performance but raise computational costs.}
To see how the number of sections used for representing each document impacts performance, we evaluate document retrieval on the InfoSeek dataset by varying the sections per document during training. As shown in~\Cref{fig:num_sec}, incorporating more sections improves MRR@10 from 7.5 to 15.7 due to leveraging richer multimodal and contextual information. However, this comes at the cost of increased computational requirements, as processing more sections raises GPU memory consumption. 

\paragraph{BCE loss is the most effective to train the section reranker.}
In our reranker design, we use a binary cross-entropy (BCE) loss by concatenating the query with each document section individually (Section + BCE), allowing the model to directly assess query-section relevance. As an alternative, we also explore a contrastive loss (Contrastive), which models section reranking similarly to document retrieval but uses sections as the retrieval units, and a variant of BCE loss (Document + BCE), where the query is concatenated with multiple sections (both positive and negative) from the same document. As shown in~\Cref{tab:rerank_design}, the Section + BCE reranker outperforms both alternatives. Specifically, contrastive loss performs the worst, suggesting that direct concatenation of query and section provides clearer relevance signals, consistent with conventional reranking approaches. Moreover, while Document + BCE leverages inter-section context, its performance might be hindered by training constraints as the model processes fewer sections during training~\citep{Jiang2024longrag,Lee2024longcontextlm}, and addressing it would be interesting future work.

\paragraph{Sections from the same document act as effective negatives to enhance reranker performance.}
In training the reranker, we investigate whether considering sections from the same document as negative examples (called In-document) is effective than other strategies, such as Top-K negatives (top-K retrieved sections based on their similarity with the input query) and In-batch negatives (positive sections from other samples in the same batch). As shown in~\Cref{tab:hard_neg}, we observe that the In-document approach achieves superior performance especially on R@1, demonstrating its ability to effectively identify the most pertinent section among highly similar sections within the same document, i.e., its training objective can encourage the reranker to focus on fine-grained distinctions between closely related sections (within the same document). 

\paragraph{Our \ours is Versatile with Different VLMs.}
To ensure the effectiveness and robustness of \ours across VLMs, we evalulate its performance with another VLM,  LLaVA-OneVision~\citep{Li2024llavaov}, with 0.5 billion parameters, in addition to LLaVA-NeXT-Interleave~\citep{Li2024llavanextinterleave} used in our main experiments. Results in~\Cref{tab:llava_ov_doc_format} show that ours continues to outperform baselines, achieving a notable 30.1\% gain in R@1 over the best baseline.

\section{Conclusion}
In this paper, we introduced \ours, a novel IR framework designed to address the limitations of conventional methods that rely on textual content of documents and their segmented passages. Our approach sits on top of recent VLMs, which enables integration and representation of diverse multimodal content (including text, images, and tables) into a unified document representation. Also, unlike prior strategies that segment documents at the passage level, our method merges these segments to maintain the document's structural coherence, while further introducing a reranking strategy for precise identification of relevant sections. Extensive experiments across various IR datasets show that \ours consistently outperforms baselines, confirming the value of interleaved multimodal representation for both document and section retrieval. We believe \ours represents a crucial step toward more comprehensive and contextually aware IR systems, capable of handling the increasing multimodality of modern information sources.
\section*{Limitations}
Due to the constraints of a single H100 GPU that we have, we represent documents by sampling a limited number of sections and averaging their corresponding embeddings (See \Cref{fig:num_sec}). While this reduces the computational demands, our findings suggest that capturing a broader document context leads to improved retrieval performance. Hence, leveraging the long context window of LVLMs with a greater number of sections could further enhance document retrieval by capturing more comprehensive information within the full document. Additionally, while using the basis model size of 0.5B (or less than 1B) parameters is a standard practice in IR literature, scaling up the basis VLMs remains an avenue for future work; however, although larger models can yield performance gains, they come at the cost of increased computational requirements. Moreover, our reranker design follows the conventional approach of concatenating the input query with individual sections. However, we believe providing the reranker with all sections together would allow the model to better leverage the contextual information from the entire document, potentially resulting in improved performance, and we leave explorations on this for future work.

\section*{Ethics Considerations}
In this work, we use a publicly available retrieval corpus for information retrieval tasks. However, the retrieval corpus may contain private, harmful, or biased content. Such undesirable features could unintentionally be reflected in the behavior of retrievers and rerankers trained on this data, potentially leading to ethical concerns during real-world deployment. However, current information retrieval techniques, including ours, do not address the retrieval of undesirable content. We recognize the critical need for safeguards to mitigate this issue. This is essential to ensure that information retrieval systems are reliable, fair, and safe for deployment.

\section*{Acknowledgements}
This work was supported by the National Research Foundation of Korea (NRF) grant funded by the Korea government (MSIT) (No. RS-2023-00256259), the grant of the Korea Machine Learning Ledger Orchestration for Drug Discovery Project (K-MELLODDY) funded by the Ministry of Health \& Welfare and Ministry of Science and ICT, Republic of Korea (grant number: RS-2024-12345678), the InnoCORE program of the Ministry of Science and ICT (N10250156), and the Institute for Information \& communications Technology Planning \& Evaluation (IITP) grant funded by the Korea government (MSIT) (RS-2019-II190075, Artificial Intelligence Graduate School Program (KAIST)).

\bibliography{custom}
\clearpage
\appendix

\begin{table*}[t]
\centering
\tiny
\resizebox{\textwidth}{!}{
    \renewcommand{\arraystretch}{1.25}
    \renewcommand{\tabcolsep}{4.5pt}
    \begin{tabular}{l c c c c c c c c c}
    \toprule
    Dataset & Query Modality & Target & Domain & Entities & Section ID & Train & Eval & Test & Corpus size \\ 
    \midrule
    Encyclopedic-VQA & Text, Text-Image & Text & Species, Landmarks & 17k & $\circ$ & 177k & 2.2k & 3.8k & 100k \\
    InfoSeek & Text-Image & Text & Diverse & 11k & $\times$ & 209k & 23k & 74k & 500k \\
    ViQuAE & Text, Text-Image & Text & Human & 1k & $\circ$ & 1.2k & 1.2k & 1.2k & 100k \\
    Open-WikiTable & Text & Table & Table & - & $\circ$ & 3.3k & 0.4k & 0.4k & 1.8k \\
    \bottomrule
    \end{tabular}
    }
\caption{Information retrieval datasets summary.}
\label{tab:dataset_stats}
\end{table*}

\section{Details of Experimental Setups \label{appendix:Implementation Details}}

\paragraph{Dataset configuration}
\Cref{tab:dataset_stats} summarizes the key properties of the datasets used in our experiment, including query modality, target item, entity domain, number of entities, and whether a section ID is provided to indicate the section containing the answer. Additionally, we provide the number of samples in the training, evaluation, and test splits, as well as the size of the corpus. We provide a more detailed explanation of the datasets below.

\begin{itemize}[itemsep=2mm, parsep=1pt, leftmargin=*]
\item \textbf{Encyclopedic-VQA}~\citep{Mensink2023encyc} is a large-scale visual question-answering (VQA) benchmark dataset, widely used for measuring the performance of multimodal IR models. Each query is linked to a specific section of a Wikipedia document (containing an answer for it) and is manually annotated by humans. Also, this dataset offers both text-only and multimodal queries. In addition to this, the queries are related to fine-grained properties of species and landmarks. Our experiments focus on the single-hop category where questions can be answered in a single retrieval step.

\item \textbf{InfoSeek}~\citep{chen2023infoseek} is a dataset designed for knowledge-intensive VQA, covering a wide range of entities (such as landmarks, animals, and food). Questions are generated by filling human-written templates with knowledge triples (subject, relation, object) available from Wikidata, which involve only the multimodal queries. As the test dataset is not available, we use the validation set as our test set, and split the training set into training and validation subsets with a 9:1 ratio.

\item \textbf{ViQuAE}~\citep{Lerner2022viquae} is a dataset focused about human entities. It provides both textual and multimodal queries, with each query linked to a specific section of a Wikipedia document that contains an answer annotated by humans, which makes it an ideal benchmark for section retrieval.

\item \textbf{Open-WikiTable}~\citep{Kweon2023openwikitable} is an extension of WikiSQL~\citep{Zhong2017wikisql} and WikiTableQuestions~\citep{Pasupat2015wikitablequestions}, designed for open-domain table question answering that requires retrieval of the most relevant table from a broader corpus. For our experiments, we adapt the WikiTableQuestions subset of Open-WikiTable, aiming at identifying the document or document section containing the target table.

\end{itemize}

\paragraph{Dataset pre-processing} In our study, we leverage interleaved multimodal content from Wikipedia documents. However, existing corpora associated with IR datasets often lack this content, typically only including the first few words of each document. Therefore, we download the HTML file of each Wikipedia document for corpus augmentation.

If the dataset provides Wikipedia URLs for its corpus, we use them to download the HTML files. Alternatively, if only entity names are provided, we generate Wikipedia URLs using those names. If a Wikipedia URL is deprecated, we remove the corresponding document from the corpus along with any associated queries. From the HTML files, we extract text, image URLs, and tables. We then split the contents by subtitles in the document where each chunk corresponds to a section. For the images, we use the image URLs to download the corresponding images, removing any invalid URLs. This process produces a dictionary that organizes text, images, and tables by section.

Since downloading contents for all documents across datasets is time- and memory-intensive, we preprocess subsets of each corpus, including documents relevant to queries in the training, evaluation, and test splits, along with unrelated documents.

\paragraph{Implementation Details}
To take advantage of larger batch sizes (while reducing GPU memory usage), we apply LoRA~\citep{Hu2022lora}. Also, to further optimize the GPU usage, we scale each image down to half of its original height and width and then combine four scaled-down images into a single composite image. All experiments are conducted using a single H100 GPU.

\section{Multi-modality Statistics in Documents}
We calculate the statistics related to multi-modality in Wikipedia documents, and find that both images and tables are evenly distributed across the whole documents. To be specific, except for the first section of documents, which contains 1.2 images on average, the distribution of images is consistent across the other sections, containing an average of 0.27 images per section. Also, tables appear less frequently, averaging 0.23 per section, but they are uniformly distributed across all sections.

\section{Efficiency of \ours}
During the retrieval process, the computational efficiency (\textit{i.e.}, the retrieval latency) of our approach remains the same regardless of the number of interleaved modalities and their compositions, as each document representation (averaged from its section embeddings) is encoded into a fixed-sized vector, whose size is the same as the case where we encode only the text. Also, even if we consider the efficiency within the document embedding process (which is typically not a concern for IR tasks as it can be done offline in parallel), the computational costs and memory usage when embedding multimodal documents are similar to the case of embedding text-only documents, as the factors that impact efficiency are not the number of multimodal content but the number of tokens within documents.

\section{Additional Experimental Results\label{appendix:add_results}}

\paragraph{Data Requirements for Models}
\begin{figure*}[t]
    \small
    \centering
    \begin{minipage}{0.48\textwidth}
        \centering
        \includegraphics[width=\linewidth]{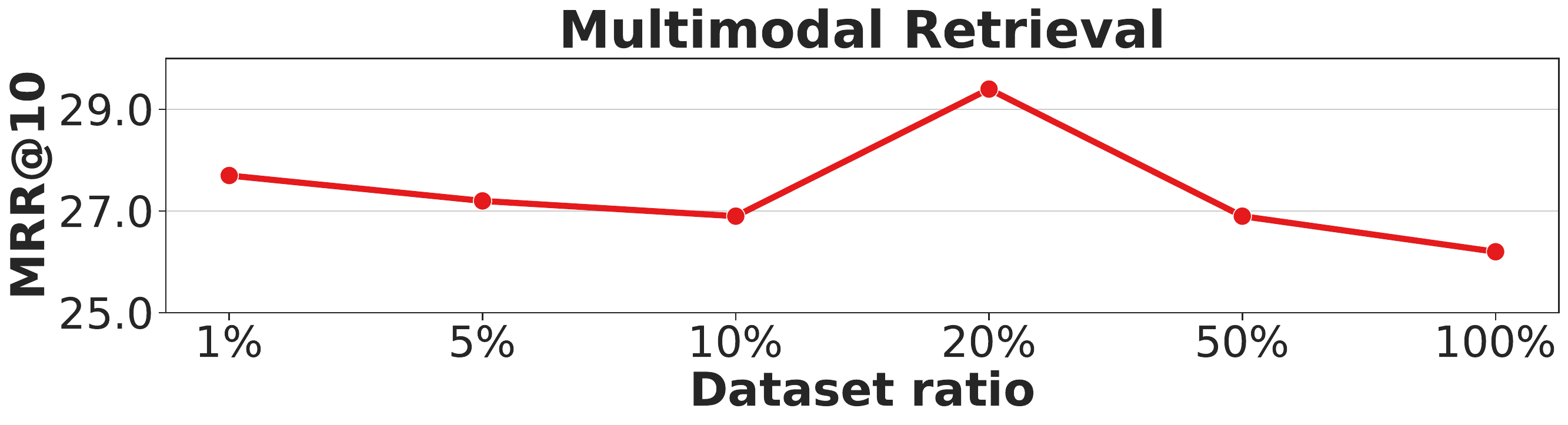}\\
        \vspace{0.02in}
        \includegraphics[width=\linewidth]{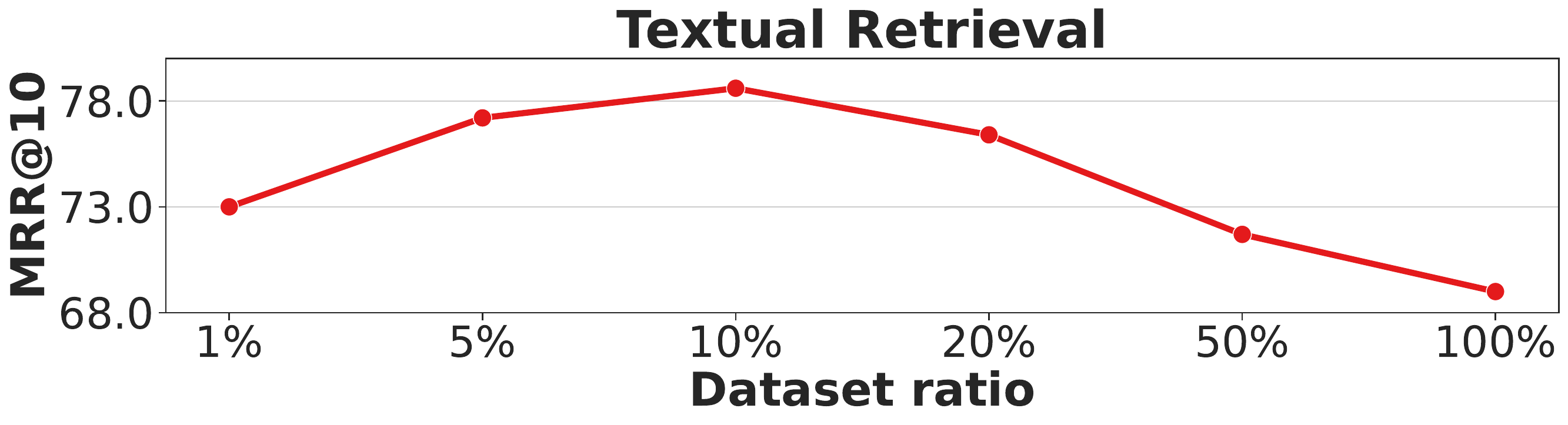}\\
        \small{\textbf{(a) Retriever performance}}
    \end{minipage}
    \hspace{0.05in}
    \begin{minipage}{0.48\textwidth}
        \centering
        \includegraphics[width=\linewidth]{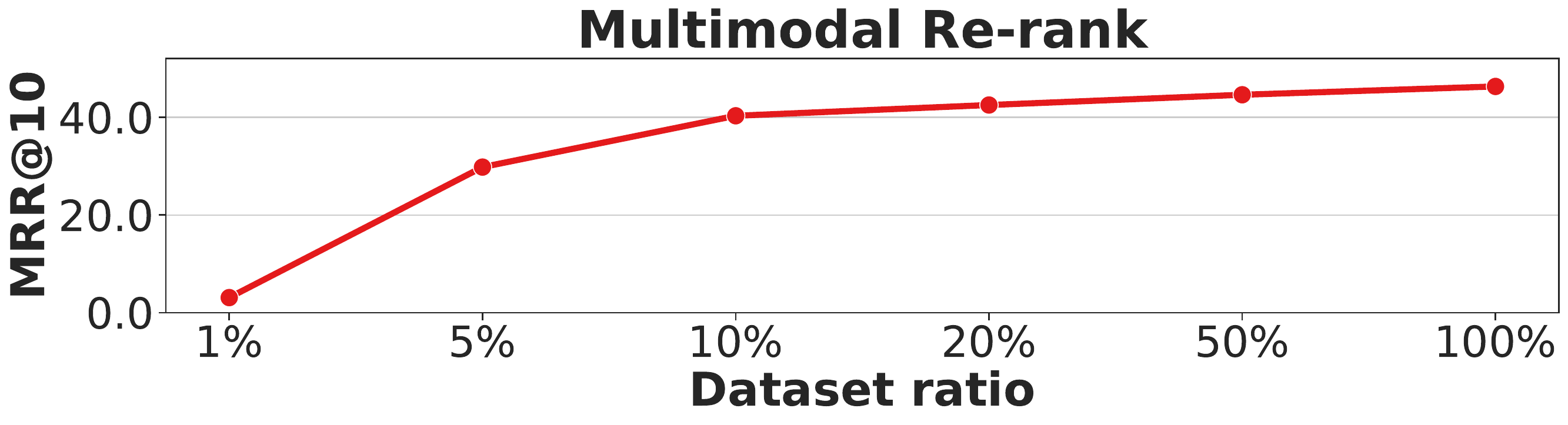}\\
        \vspace{0.02in}
        \includegraphics[width=\linewidth]{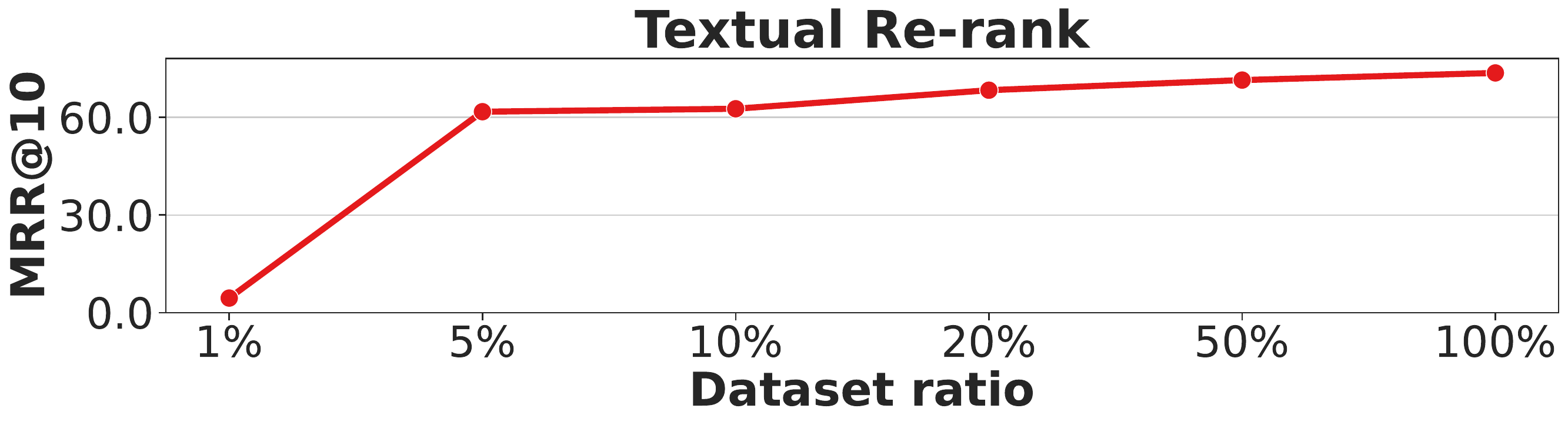}\\
        \small{\textbf{(b) Reranker performance}}
    \end{minipage}
\caption{ Retrieval performance with different dataset sizes for training. \textbf{(a)}: When training a retriever, large datasets rather deteriorate the retrieval performance as it may be overfitted, resulting in low generalization. \textbf{(b)}: On the other hand, a larger dataset size is beneficial to training a re-ranker.}
\label{fig:num_dataset}
\end{figure*}
We analyze the effect of different dataset sizes for training on retriever and reranker performance. To achieve this, we randomly prune samples in the Encyclopedic-VQA dataset at various ratios and report the performance of models trained on these subsets. In~\Cref{fig:num_dataset} \highlight{(a)}, we observe that too many samples can degrade retrieval performance. Also, retrieval of textual queries requires fewer samples to reach its optimal performance compared to multimodal retrieval. Similarly, in~\Cref{fig:num_dataset} \highlight{(b)}, section retrieval for multimodal queries requires 10\% of the dataset to achieve 80\% of the full-dataset performance, while section retrieval for textual queries needs only 5\%. These observations suggest that additional modalities increase the need for more data. This accounts for the inferior performance of the interleaved format in the ViQuAE experiments (\Cref{tab:information_retrieval} \highlight{Right}). The ViQuAE dataset, at only 2.2\% of the size of Encyclopedic-VQA, may be small for the reranker to effectively learn multimodal query-section alignments. We also observe that section retrieval is more challenging, with more samples improving the reranker's performance. This explains why the ViQuAE reranker has much lower section retrieval scores compared to the one trained on the Encyclopedic-VQA (\Cref{tab:information_retrieval} \highlight{Right}). Given the challenge of obtaining large query-section pair samples, exploring more effective reranker training pipelines is necessary.

\section{Case Study\label{appendix:case studies}}
\begin{figure*}[t]
    \centering
    \begin{minipage}{\linewidth}
        \begin{tcolorbox}[colback=gray!5!white,colframe=black]
            Q: What is the name of the park on the north shore of foster reservoir?
        \end{tcolorbox}
        \centering
        \includegraphics[width=1.0\linewidth]{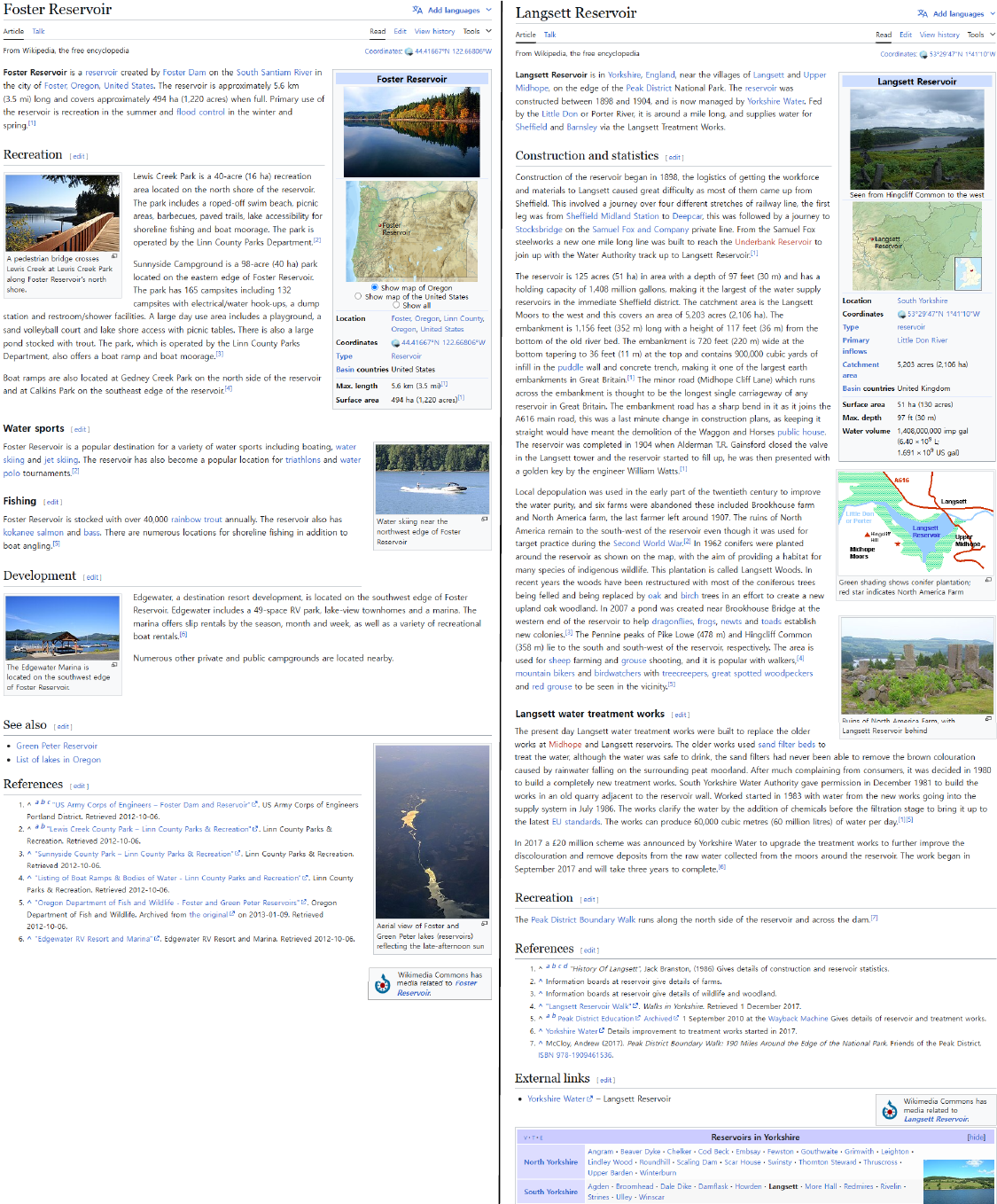} \\
        \par
        \vspace{0.05in}
        \raggedright{\textbf{(a) Interleaved Multimodal Document Retrieval \phantom{,,,,,,,,,,,,,} (b) Text-only Document Retrieval}}
        \par
        \vspace{-0.05in}
        \caption{Retrieved documents across different document formats for document retrieval with a given textual query. \textbf{(a)}: A document retrieved when represented leveraging interleaved multimodal contents within documents (ours). \textbf{(b)}: A document retrieved when using only textual format}
        \label{fig:case_study_text}
    \end{minipage}
\end{figure*}

\begin{figure*}[t]
    \centering
    \begin{minipage}{\linewidth}
        \begin{tcolorbox}[colback=gray!5!white,colframe=black]
            \includegraphics[width=2.5cm]{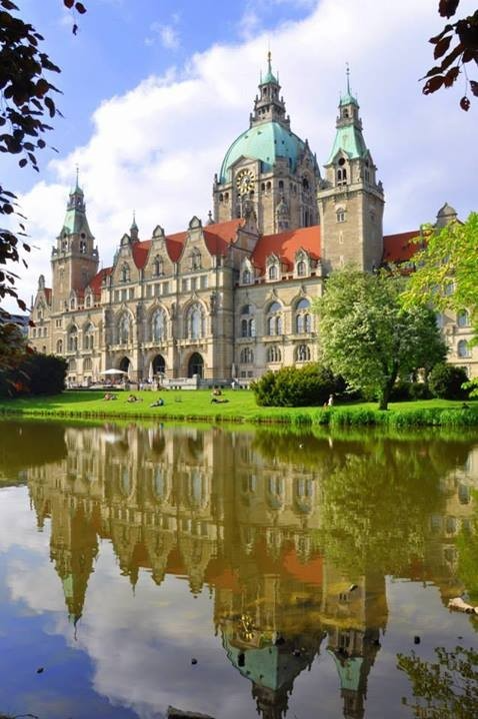}
            \hspace{0.1in}Q: Who designed this building?
        \end{tcolorbox}
        \centering
        \includegraphics[width=1.0\linewidth]{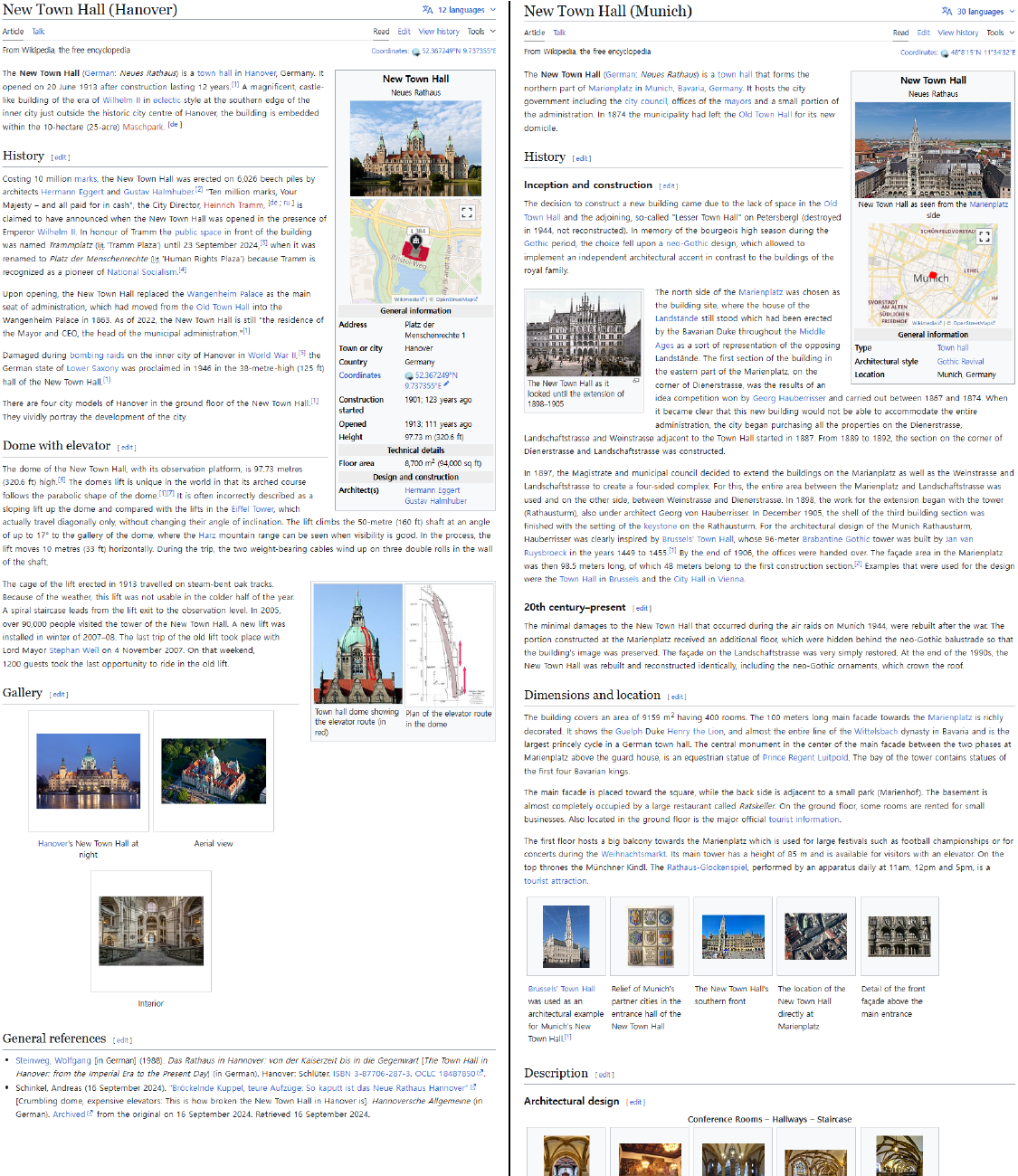}
        \par
        \vspace{0.05in}
        \raggedright{\textbf{(a) Interleaved Multimodal Document Retrieval \phantom{,,,,,,,,,,,,,} (b) Text-only Document Retrieval}}
        \par
        \vspace{-0.05in}
        \caption{Retrieved documents across different document formats for document retrieval with a given multimodal query. \textbf{(a)}: A document retrieved when represented leveraging interleaved multimodal contents within documents (ours). \textbf{(b)}: A document retrieved when using only textual format}
        \label{fig:case_study_multimodal}
    \end{minipage}
\end{figure*}

We conduct case studies to demonstrate the advantages of our approach in document retrieval with textual and multimodal queries. In~\Cref{fig:case_study_text} and~\Cref{fig:case_study_multimodal}, we illustrate the instances where our approach, which leverages interleaved multimodal contents (e.g., images, tables, and text) within documents, retrieved correct documents for given queries, while the conventional one, which represents documents using only textual data, retrieved documents that appeared to be relevant but were not actually related to the queries.

In~\Cref{fig:case_study_text}, a textual query asks for the name of the park located on the north shore of Foster Reservoir. The conventional approach retrieved a document containing unrelated information about a different reservoir. While this document includes terms such as "Peak District National Park" and "North America farm," which make the document superficially relevant, it fails to answer the query. In contrast, our approach identified the document containing the correct answer to the given query.

The advantages of integrating multimodal content into document representation become more apparent in document retrieval with multimodal queries, as shown in~\Cref{fig:case_study_multimodal}. For a query consisting of an image of a town hall in Hanover and a textual question about its designer, both our approach and the conventional one retrieved documents about town halls in Germany. However, our approach pinpointed the exact document about the town hall in Hanover, indicating that Hermann Eggert designed the building. The conventional method retrieved a document about a town hall in Munich, which is somewhat related but not an exact match to the query image or question.

These cases underscore the benefits of leveraging multimodal content in information retrieval. Integrating interleaved multimodal elements, our approach aligns more effectively with the input query, resulting in more accurate and fine-grained retrieval. This superiority is supported by \citet{Xu2024rir}, which highlights that models perform better when prompted with rich multimodal information, enabling them to capture alignments across modalities and enhance the representation of given inputs.


\end{document}